\documentclass[journal]{IEEEtran}
\usepackage{cite}
\usepackage{color,xcolor}
\usepackage{amsmath,amssymb,amsfonts}
\usepackage{hyperref}
\usepackage{textcomp}
\usepackage{graphics}
\usepackage{graphicx}
\usepackage{subfigure}
\usepackage{setspace}
\usepackage{booktabs}
\usepackage{makecell}
\usepackage{threeparttable}

\begin{document}
%
\title{MRFI: An Open Source Multi-Resolution Fault Injection Framework for Neural Network Processing}

\author{
Haitong Huang, 
Cheng Liu,~\IEEEmembership{Member,~IEEE}, 
Bo Liu,
Xinghua Xue, 
\\
Huawei Li,~\IEEEmembership{Senior Member,~IEEE}, 
Xiaowei Li,~\IEEEmembership{Senior Member,~IEEE}
\thanks{This work was supported in part by the National Natural Science Foundation of China under Grant No. 62174162 and No. 62090024, and the Space Trusted Computing and Electronic Information Technology Laboratory of Beijing Institute of Control Engineering (BICE) under Grant OBCandETL-2022-07. \textit{(Corresponding author: Cheng Liu.)}}
\thanks{Haitong Huang, Cheng Liu, Xinghua Xue, Huawei Li, and Xiaowei Li are with both State Key Lab of Processors, Institute of Computing Technology, Chinese Academy of Sciences, Beijing 100190, China, and University of Chinese Academy of Sciences, Beijing 100190, China. (e-mail: \{huanghaitong21s, liucheng\}@ict.ac.cn)}
\thanks{Bo Liu is with Beijing Institite of Control Engineering, Beijing 100190, China.}
}

\markboth{IEEE TRANSACTIONS ON VERY LARGE SCALE INTEGRATION (VLSI) SYSTEMS, VOL. XX, NO. XX, XXX 2023}{Haitong Huang \MakeLowercase{\textit{et al.}}: MRFI: An Open Source Multi-Resolution Fault Injection Framework for Neural Network Processing}

\maketitle

\begin{abstract}

To ensure resilient neural network processing on even unreliable hardware, comprehensive reliability analysis against various hardware faults is generally required before the deep neural network models are deployed, and efficient error injection tools are highly demanded. However, most existing fault injection tools remain rather limited to basic fault injection to neurons and fail to provide fine-grained vulnerability analysis capability. In addition, many of the fault injection tools still need to change the neural network models and make the fault injection closely coupled with normal neural network processing, which further complicates the use of the fault injection tools and slows down the fault simulation. Due to the various fault injection implementations and error metrics, it makes the comparison between different fault-tolerant study difficult. To this end, we propose MRFI, a highly configurable multi-resolution fault injection tool for deep neural networks. It enables users to modify an independent fault configuration file rather than neural network models for the fault injection and vulnerability analysis. Particularly, it integrates extensive fault analysis functionalities from different perspectives and enables multi-resolution investigation of the vulnerability of neural networks. In addition, it does not modify the major neural network computing framework of PyTorch. Hence, it allows parallel processing on GPUs naturally and exhibits fast fault simulation according to our experiments. Moreover, we also have the fault injection calibrated with fault simulation with architectural details and validate the accuracy of the proposed fault injection. Finally, MRFI is also open sourced on Github\footnote{MRFI \url{https://github.com/fffasttime/MRFI}}.


\end{abstract}


\IEEEpeerreviewmaketitle

\section{Introduction}
Deep neural network models are increasingly used in safety-critical applications like autonomous driving and avionics, but computational errors caused by imperfect hardware such as process variations, defects and impurities in materials, and noise can lead to unexpected inference and cause severe consequences in these safety-critical applications  \cite{mittal2020survey} \cite{ibrahim2020soft} \cite{wei2023tc} \cite{nguyen2023craft}. Even for the non-critical applications, reliability may also be of vital importance when the overall system scales out in the large language model era \cite{wang2023reliable} \cite{he2023understanding} \cite{wu2023transom}. To guarantee resilient deep learning, it is crucial to thoroughly evaluate the reliability of neural network models before their deployment. While it is prohibitively expensive to perform fault injection to realistic chips, simulation based error injection tools such as PyTorchFI \cite{PytorchFI}, Ares \cite{Ares2018}, TensorFI \cite{TensorFI}, TorchFI \cite{torchfi2019}, and MindFI \cite{Zhen2021MindFI} have been extensively developed. These tools typically provide basic functionalities such as error injection and reliability analysis to general neural network processing and significantly alleviate the design efforts required for neural network reliability analysis in presence of various hardware errors. 

Nevertheless, it has become evident that existing fault injection tools are insufficient to address complex neural network reliability analysis for the following aspects. First, they generally utilize hardware fault models such as bit flip model to simulate the influence of hardware errors on neural network processing, but they do not model any specific hardware architectures and the correlation between the neuron-wise error injection and realistic hardware-based fault injection remains unclear \cite{PytorchFI} \cite{Ares2018} \cite{TensorFI} \cite{xue2023exploring} \cite{xue2023soft} \cite{Mahmoud2021ISSRE}. The accuracy of the fault simulation needs to be calibrated such that it can be utilized as a common metric to reliability study of neural network processing. Second, there is a lack of standard error injection metrics and neural network evaluation metrics that can be utilized to align the fault simulation results and distinct fault simulation results may be obtained from these different fault injection tools, which dramatically affects the reliability study of neural network processing. For instance, many prior fault injection tools fail to clarify the model quantization during reliability evaluation and biased evaluation results are obtained because of the quantization setups \cite{statistical2023}. Third, many of the fault injection tools are closely coupled with the neural network frameworks such as PyTorch. The error model and error injection may vary with the different neural network layers \cite{FIdelity}, which also complicates the fault simulation. In addition, fine-grained selective fault-tolerant design \cite{libano2018selective} \cite{ahmadilivani2023enhancing} \cite{SNR2021} has also been explored and it poses great challenges to the fault simulation and evaluation of existing fault injection tools. These unique fault simulation and evaluation requirements further make the fault injection frameworks difficult to scale and port to GPUs or other parallel computing systems.

To address the above problems, we present MRFI, a highly configurable open-sourced fault injection tool for reliability investigation and evaluation of neural network processing. On the one hand, it provides unified error models, neural network quantization setups, and error injection mechanisms to ensure consistent metrics of error injection and reliability evaluation. Moreover, it is compared to typical fault injection platforms with more detailed architectural details and calibrated for more fair comparison. On the other hand, it provides a multi-resolution vulnerability investigation of neural network processing from distinct angles and granularities, which can be utilized to guide selective fault-tolerant neural network design conveniently. In addition, MRFI utilizes a tree structure for flexible error injection configurations and reliability evaluation of different neural network modules without modifying the neural network processing engines, which will not affect the underlying parallel computing engines on GPUs and can achieve the state-of-the-art fault simulation.

The contributions of this paper can be summarized as follows:
\begin{itemize}
\item We present a multi-resolution fault injection framework MRFI. It enables vulnerability analysis of neural network models with different granularities, which can be utilized to guide selective protection of neural network processing from distinct perspectives. In addition, we have it calibrated with architecture-specific fault injection frameworks and validates its fault simulation accuracy. 
\item MRFI provides unified error injection and reliability evaluation with flexible simulation configurations. Particularly, it allows fine-grained vulnerability analysis from different angles and can be utilized for selective protection against various hardware errors. 
\item MRFI is built on top of PyTorch and does not affect the computing engine of PyTorch. Hence, the fault injection and reliability evaluation can be scaled to parallel computing systems like GPUs conveniently and achieve competitive simulation speed.

\item MRFI is open sourced on github and has been released on PYPI and can be directly installed using \texttt{pip install mrfi}. We also conducted sufficient unit testing on the fault injection code to ensure its correctness.
\end{itemize}

\begin{table}[htbp]
  \begin{threeparttable}
  \centering
  \caption{Comparison of different fault injection frameworks including MRFI, PyTorchFI\cite{PytorchFI}, and Ares\cite{Ares2018}}
    \begin{tabular}{cccc}
    \toprule
     &MRFI &PyTorchFI &Ares \\
    \midrule
    \multicolumn{4}{c}{{activaion injection}} \\
    \hline
    permanent fault&\checkmark&\checkmark&× \\
    transient fault&\checkmark&\checkmark&\checkmark \\
    \hline
    \multicolumn{4}{c}{{weight injection}} \\
    \hline
    permanent fault&\checkmark&\checkmark&\checkmark \\
    transient fault&\checkmark&×&× \\
    \hline
    \multicolumn{4}{c}{{number of faults}} \\
    \hline
    fixed number fault&\checkmark&\checkmark&× \\
    random error rate&\checkmark&×&\checkmark \\
    \hline
    \multicolumn{4}{c}{{fault injection on quantized data}} \\
    \hline
    quantized fault&\checkmark&\checkmark&\checkmark \\
    full quantization&\checkmark&×&\checkmark \\
    \makecell{quantization\\ method selection}&\checkmark&×&× \\
    \makecell{quantization \\ parameter settings}&\checkmark&limited&limited \\
    \hline
    \multicolumn{4}{c}{{fault injection on floating point data}} \\
    \hline
    fixed value inject&\checkmark&\checkmark&× \\
    random perturbation&\checkmark&×&\checkmark \\
    random bit flip&\checkmark&×&× \\
    \hline
    \multicolumn{4}{c}{{internal observers}} \\
    \hline
    clean activations&\checkmark&×&× \\
    fault impact&\checkmark&×&× \\
    \hline
    \multicolumn{4}{c}{{functional flexibility}} \\
    \hline
    custom error model&\checkmark&\checkmark&× \\
    \makecell{multi-resolution\\ evaluation}&B,N,C,L\tnote{1}&N,L&L \\
    \hline
    \multicolumn{4}{c}{{fine-grained simulation configuration}} \\
    \hline
    by python program&\checkmark&limited&× \\
    by configuration file&\checkmark&×&×\\
    \bottomrule
    \end{tabular}%
  \label{tab:feature_comparesion}%
  \begin{tablenotes}
    \item[1] B: bit, N: neuron, C: channel, L: layer
  \end{tablenotes}
  \end{threeparttable}
\end{table}%

\section{Related Work}

Research on the reliability and fault tolerance of deep learning revolves around error simulation, fault evaluation, fault-tolerant model design, and fault-tolerant hardware design. Error simulation methods can be divided into hardware-level simulation and model-level simulation. Hardware-level simulation is generally more accurate but has limited generality and performance due to the need to simulate the underlying logic of specific hardware architectures. Notable works in this area include NVIDIA SASSIFI for fault injection simulation on GPUs\cite{SASSIFI} \cite{tsai2021nvbitfi} and fault simulation on FPGAs or ASICs\cite{libano2018selective} \cite{gambardella2019efficient} \cite{xu2020persistent} \cite{xu2021reliability} \cite{tan2023saca}. These error simulation methods often require tens to thousands of times the normal execution time.

On the other hand, model-level evaluation and simulation have better performance, making them widely studied in previous works. Some notable works in this area include \cite{PytorchFI} \cite{Ares2018} \cite{TensorFI} \cite{torchfi2019} \cite{Zhen2021MindFI} \cite{Li2017understanding}. Some works have shown that accurate hardware-level reliability results can be obtained through model-level simulation. For example, in Ares\cite{Ares2018}, the authors obtained consistent measurements with silicon measurement by injecting errors at the model level. Yi He et al. \cite{FIdelity} mapped hardware unit error rates to model neurons to obtain accurate model-level evaluation results. Xinghua Xue et al. \cite{xue2022winograd} improved the accuracy of computational evaluation by focusing on operations. Some works aim to improve the performance of error simulation. For example, PyTorchFI\cite{PytorchFI} optimized the implementation of fault injection simulation, reducing the additional time overhead of single-point fault injection to a negligible level. The closest state-of-the-art tools to our work are PyTorchFI and Ares, which offer more comprehensive functionality compared to other works. \autoref{tab:feature_comparesion} shows a summary of their features.

For fault evaluation, recent works have proposed improved statistical metrics and mathematical modeling analysis to further reduce the number of error simulations required. In HarDNN\cite{Mahmoud2020HarDNNFM}, the authors explored various surrogate test metrics to replace simple model accuracy measurements, resulting in a 10-times reduction in simulation count under the same simulation accuracy. BinFI \cite{chen2019BinFI} models error propagation and introduces binary search to reduce the number of simulations. \cite{statistical2023} achieved a two-order-of-magnitude speedup in fault evaluation by using statistical modeling. Previous fault injection frameworks did not consider these optimizations, whereas our framework's flexibility allows researchers to conveniently perform these evaluations.

After fault evaluation, safety-critical systems often require fault-tolerant design. Fault-tolerant design can be applied at both the software and hardware levels, such as adjusting models for fault tolerance or introducing additional hardware redundancy computations \cite{liu2022special} \cite{li2020ftt} \cite{ning2021ftt} \cite{zhang2018analyzing} \cite{reagen2016minerva}. These solutions include Selective Hardening \cite{libano2018selective}, Algorithm-Based Fault Tolerance \cite{hari22022abft}, adjustment of quantization parameter settings \cite{SNR2021}, adjustment of activation layer values to limit numerical magnitude \cite{fitact2022}, fault-tolerant retraining \cite{xu2021r2f}, and more. Evaluating these solutions requires a more flexible fault injection framework.

\section{MRFI Framework}

\begin{figure*}[htbp]
    \centering
    \includegraphics[width=0.95\linewidth]{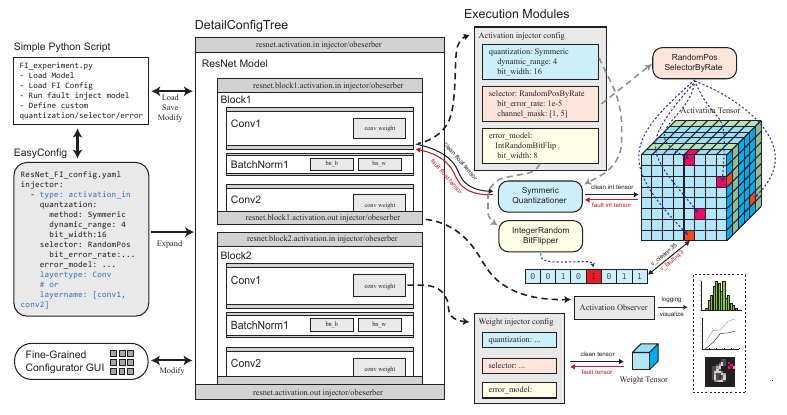}
    \caption{The MRFI tool integrates multiple modules. The configuration section can be used for simple or detailed experiments. EasyConfig provides a unified annotation error configuration for multiple layers, while DetailConfigTree corresponds to each neural network module and can be modified separately. The execution module is decomposed into quantizers, selectors, and error models according to the requirements of the fault injection, and provides various predefined functions.}
    \label{fig:overallpic}
\end{figure*}

\subsection{MRFI Overview}

The overall structure of MRFI is illustrated in \autoref{fig:overallpic}. The framework as a whole can be divided into two parts: the configuration part and the execution modules, which are further divided into several functional modules.

MRFI provides two main approaches to configure different experimental requirements: EasyConfig for simple and preliminary experiments, and a detailed error injection configuration tree for more complex needs.
For simple experiments, EasyConfig allows users to specify the layers and fault injection method through a config file. This configuration applies uniformly to the specified set of layers, enabling users to conduct fault injection experiments by loading the corresponding error injection configuration, similar to working with a typical PyTorch model.
To address complex and precise error injection requirements without modifying the experimental model, MRFI employs a detailed error injection configuration tree. This tree structure corresponds to the nested structure of neural network modules in PyTorch, recording the error injection configuration of each module and layer. Users can easily adjust the parameters within the configuration tree to meet their specific needs. MRFI's error injection module then dynamically performs error injection based on the configured settings.

The execution modules of MRFI offer multiple commonly used injection methods that encompass the majority of error injection requirements for experiments. For enhanced flexibility, users can also customize the execution module according to their specific needs.

The structured configuration approach of MRFI offers several advantages over previous manual error injection implementations. In the traditional manual approach, experimenters typically need to modify both the neural network model code and the error injection parameters to accommodate different error injection requirements. This process is hard especially for modern large-scale models with numerous layers. Additionally, manually adjusting error injection parameters for multiple fine-grained settings in dozens of model layers is cumbersome, making experimental management challenging. MRFI overcomes these issues by decoupling the error injection configuration from the model definition and organizing it into separate modules, facilitating error injection experiments with specific configurations.

\subsection{Integration with PyTorch}

MRFI is built upon PyTorch, the widely adopted neural network training framework. Error injection in MRFI is achieved through PyTorch hook, allowing for seamless integration with the model. PyTorch serves as the foundation for MRFI due to its extensive user base and well-established software ecosystem. Additionally, we have found that PyTorch's dynamic graph approach aligns well with error injection scenarios that necessitate dynamic modification and configuration. The hook function in PyTorch enables dynamic observation and modification of intermediate data within network modules, making it highly suitable for fine-grained error injection experiment configuration. 
An alternative implementation approach involves adding or replacing layers in the model with fault injection operations. However, this method requires prior knowledge of the original layer implementations, which is impractical given the diverse range of user-defined modules and layers. It is worth noting that error injection at the model level only requires access to the model's weights and feature map information for modification. By leveraging hooks, rewriting of the underlying computation implementation is not necessary. Our execution module also mainly uses PyTorch tensors, which allows the error injection process to be naturally used on the CPU and GPU to maximize performance.

\subsection{Major MRFI Execution Modules}

As depicted in \autoref{fig:overallpic}, MRFI incorporates three key components for fault injection: quantizer, selector, and error model. Additionally, an observer feature is available for monitoring neuron values and their error impact within the model.

\subsubsection{Quantizer}
In MRFI, we recognize the significance of quantization methods in evaluating fault tolerance. Quantization is commonly employed during inference for fixed-point or integer computations, differing from the floating-point storage and computation used during model training. Previous studies highlight the varying fault-tolerance characteristics of different numerical representations and the influence of quantization choices on model accuracy and fault tolerance. To facilitate accurate fault simulation for quantized models, MRFI offers multiple configurable quantization options. These simulated quantizations encompass the quantization and de-quantization steps, converting tensors from floating-point to integer format and vice versa. The quantizer serves as a wrapper for the fault injection component, enabling the provision of integer tensors to the selector and error model for simulating integer computation and error occurrence on real hardware. It is worth noting that the quantizer is optional, and users can omit it if they solely require simulations with floating-point models.

\subsubsection{Selector}
The selector determines the locations within the tensor where errors are injected. By defining the selector, users can specify the manner in which errors manifest. For example, a fixed position selector can be employed to evaluate the impact of permanent errors. Conversely, a random position selector serves as a random engine to simulate the occurrence of soft errors. Soft errors are often associated with a bit error rate, and the random position selector efficiently generates erroneous positions based on this rate. Users also have the flexibility to customize filtering masks at the selector level or implement custom selectors to simulate fine-grained selective fault-tolerant protection.

\subsubsection{Error model}
The error model dictates how errors are injected into the neuron values at the selected error positions. The error model may rely on the original neuron values, such as random bit flips for integers or floating-point numbers, fixed bit flips, or the addition of perturbations following a normal distribution. Alternatively, the error model may be independent of the original neuron values, involving stuck-at 0 faults, faults with specific fixed numerical values, or random uniform integers or floating-point numbers. MRFI incorporates these common error models, while also allowing users to define custom error models conveniently.

\subsubsection{Observer}
MRFI provides internal observers to enhance the visibility of neuron values and their error impact within the model. Observers that do not require fault injection can be employed to monitor the distribution of internal activation values. For example, an observer can track the maximum and minimum values to determine the dynamic range of activations, aiding in the determination of model quantization parameters. Another type of observer enables the comparison between fault injection and non-fault injection scenarios. Before initiating fault injection, this observer conducts a golden run comparison analysis. It can utilize a counter to tally the number of neuron values affected by fault injection or employ an RMSE observer to gauge the extent of error propagation.

\subsection{Implementation details of MRFI}

In this section, we explain several key points in the implementation of MRFI, which make it simple to use and powerful in functionality.

\subsubsection{Human-readable configuration files}
MRFI employs DetailConfigTree as its internal configuration for execution, which corresponds to each PyTorch module. EasyConfig, on the other hand, can match multiple layers and is expanded into DetailConfigTree before usage. Both EasyConfig and DetailConfigTree can be loaded, saved, and modified using human-readable YAML files. Figure 1 presents an example of EasyConfig configuration, specifying the injection method, selected layer, and execution module parameters. DetailConfigTree uses similar configuration parameters but adopts a recursive structure, with each child node corresponding to a specific module or layer. For large deep models, creating and modifying DetailConfigTree manually can be challenging. Therefore, MRFI simplifies the initial stage by starting with a simple EasyConfig, parsing the PyTorch model, and performing fuzzy or precise matching on the user-selected layers. The parameters from EasyConfig are then expanded onto the DetailConfigTree to facilitate the execution.

\subsubsection{Dynamic fault injection execution}
As mentioned above, MRFI utilizes PyTorch hooks to implement runtime fault injection of model variables. During the loading phase, hooks are inserted into the target model's layers. During the experimental stage, when a hook with fault injection settings is executed, MRFI selects the corresponding execution module based on the DetailConfigTree configuration and invokes them in order. Notably, MRFI adopts a fully dynamic approach to execute configurations, allowing for dynamic modification of fault injection parameters during the experiment. This design maximizes flexibility, enabling users to easily conduct experiments with various parameters by dynamically adjusting the internal configuration within a Python for-loop script. This eliminates the need for repetitive manual modifications of the text configuration file. Therefore, MRFI surpasses previous tools that offer limited or fixed injection error settings, effectively meeting diverse experimental needs.

\subsubsection{Extensible execution modules}
MRFI divides error injection into simple parts, it is easy to customize and add new functionality. For instance, a selector function receives the shape of a target tensor and generates a list o fault injection positions, while the error model function processes the original tensor data and returns the error results based on the specified error model. As long as a function adheres to the defined interfaces, it can be written in the configuration and executed by MRFI. This extensibility enables researchers to tailor MRFI to their specific requirements and incorporate new error injection techniques with ease.

\section{Experiment Results}

In this section, we present the results of various error injection experiments conducted using MRFI. Our focus is on demonstrating experimental approaches not covered by previous error injection tools, highlighting the necessity of utilizing a more fine-grained error injection framework.

For our experiments, we primarily employ VGG and ResNet models evaluated on the ImageNet dataset. In \autoref{sec:observing}, we also utilize a smaller dataset, cifar-10, to visualize error propagation.

\subsection{Basic BER-Accuracy fault inject experiment\label{sec:basic_use}}

\begin{figure}[htbp]
    \centering
    \subfigure[VGG16 weights]{
    \includegraphics[width=0.45\linewidth]{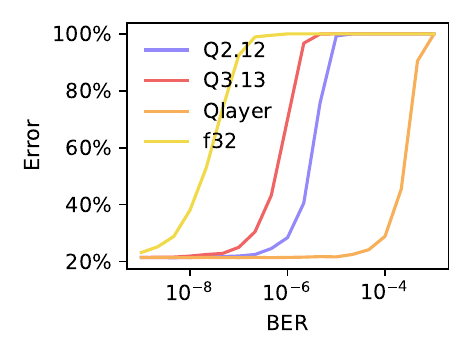}
    }
    \subfigure[ResNet18 weights]{
    \includegraphics[width=0.45\linewidth]{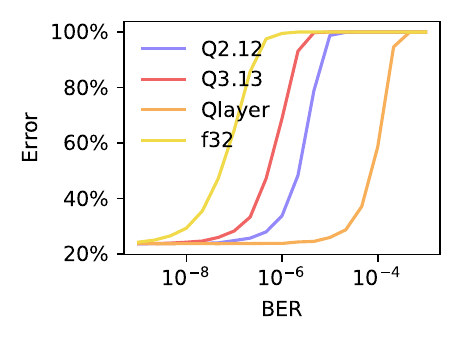}
    }
    \subfigure[VGG16 activations]{
    \includegraphics[width=0.45\linewidth]{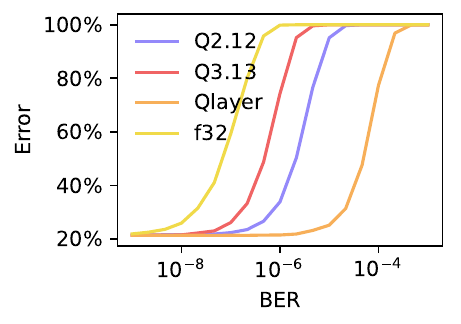}
    }
    \subfigure[ResNet18 activations]{
    \includegraphics[width=0.45\linewidth]{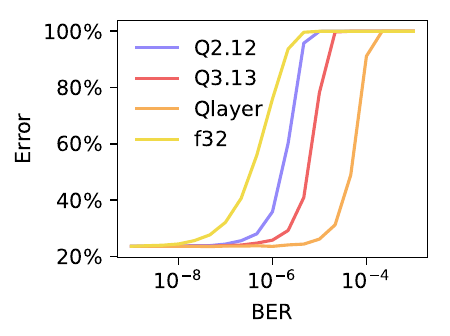}
    }
    \caption{Relation between bit error rates and classification error rate of different data type.}
    \label{fig:BER_Acc}
\end{figure}

We begin by testing and evaluating MRFI's performance under simple configurations. Specifically, we conduct random bit flipping error injection on the quantization weight of the VGG16 model using the fixed-point quantization settings of 2.12 and 3.13. Additionally, MRFI supports different data types and quantization methods. For comparison, we include the original 32-bit floating point (f32) and layer-wise range quantization (Qlayer).

As illustrated in \autoref{fig:BER_Acc}(a)(b), our results of fixed-point quantization with settings 2.12 and 3.13 align with those obtained from Ares\cite{Ares2018}, exhibiting a significant increase in bit error rates at $BER = 10^{-6}$. In contrast, floating-point types exhibit lower error tolerance, while layered quantization demonstrates superior fault tolerance, corroborating the conclusions from SNR\cite{SNR2021}. In \autoref{fig:BER_Acc}(c)(d), the injection of activation values exhibits a similar trend, but not entirely the same.

While the above experiments could be conducted separately using different fault injection tools in the past, the advantage of MRFI lies in its integration of diverse functionalities within a unified framework.

\subsection{Observing error propagation\label{sec:observing}}

To gain deeper insights into error propagation within the network, MRFI provides a straightforward operation to apply error observers to all network layers. These observers enhance transparency and offer diverse perspectives for in-depth research on error impact.

\begin{figure}[htbp]
    \centering
    \includegraphics[width=1.0\linewidth]{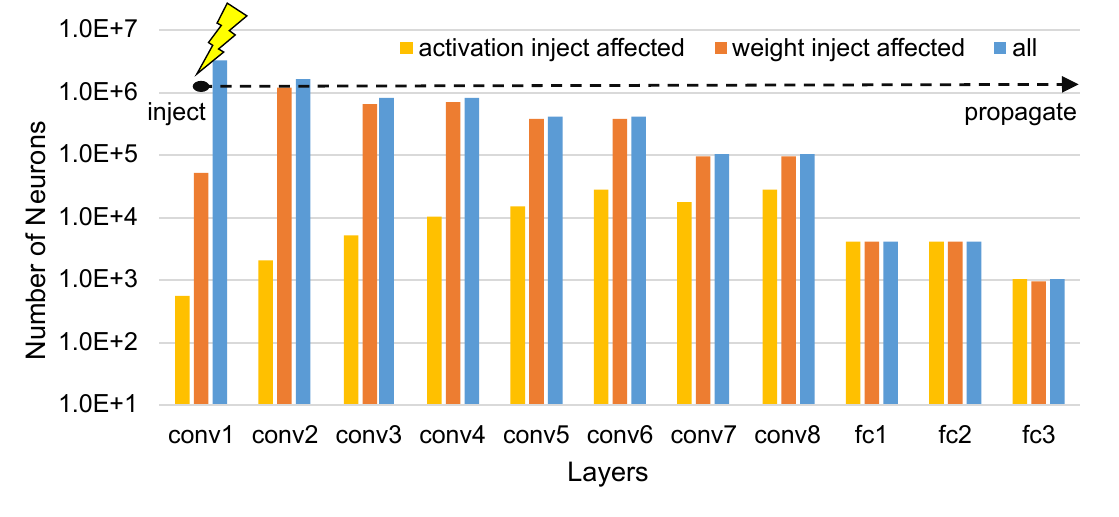}
    \caption{Number of affected neurons of one bit flip on first layer in VGG-11.}
    \label{fig:spread_number}
\end{figure}

Firstly, we employ a simple equation comparison observer to assess the number of neurons affected by a single injection. We perform one-bit flipping on the activation values and weights of the first convolution layer (conv1) of VGG11, and then calculate the average number of errors across 10,000 input images. As shown in \autoref{fig:spread_number}, the number of affected neurons increases rapidly along the layers, with nearly all neurons in the fully-connected layers displaying differences from their original values. The golden run accuracy and accuracy after activation injection both yield $76.49\%$, while the accuracy after weight injection slightly decreases to $76.34\%$. This indicates the network's resilience to errors, while also highlighting that the number of affected neurons is not a good indicator of the severity of error effects.

\begin{figure}[htbp]
    \centering
    \subfigure[MAE]{
        \includegraphics[width=0.9\linewidth]{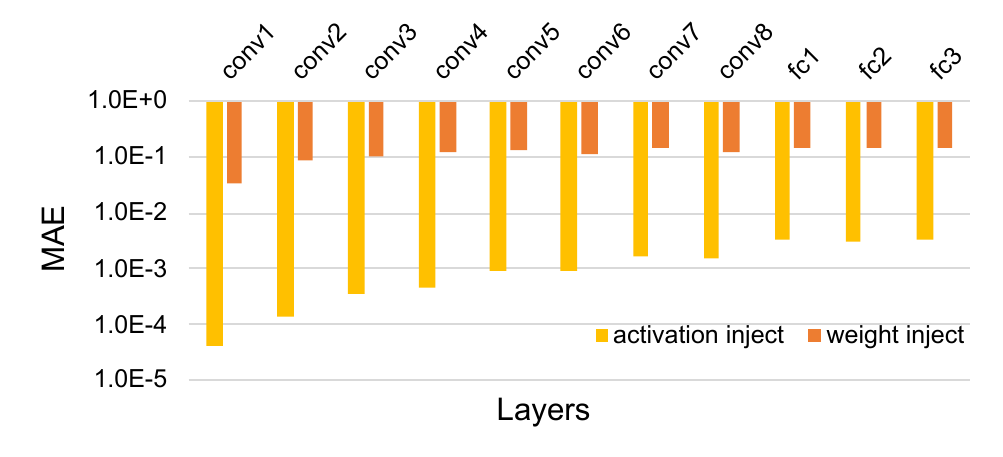}
    }
    \subfigure[RMSE]{
        \includegraphics[width=0.9\linewidth]{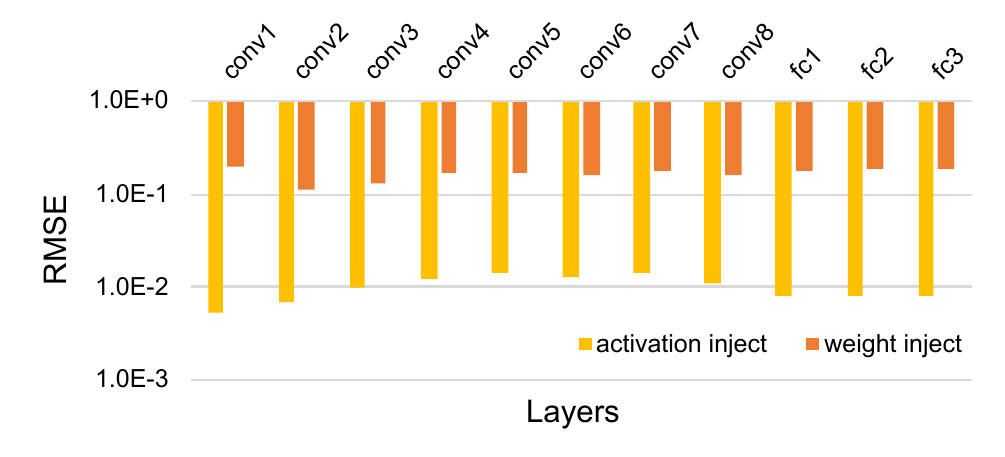}
    }
    \caption{Evaluate error propagation by better metrics.}
    \label{fig:spread_metrics}
\end{figure}

\begin{figure*}[htbp]
    \centering
    \includegraphics[width=0.75\linewidth]{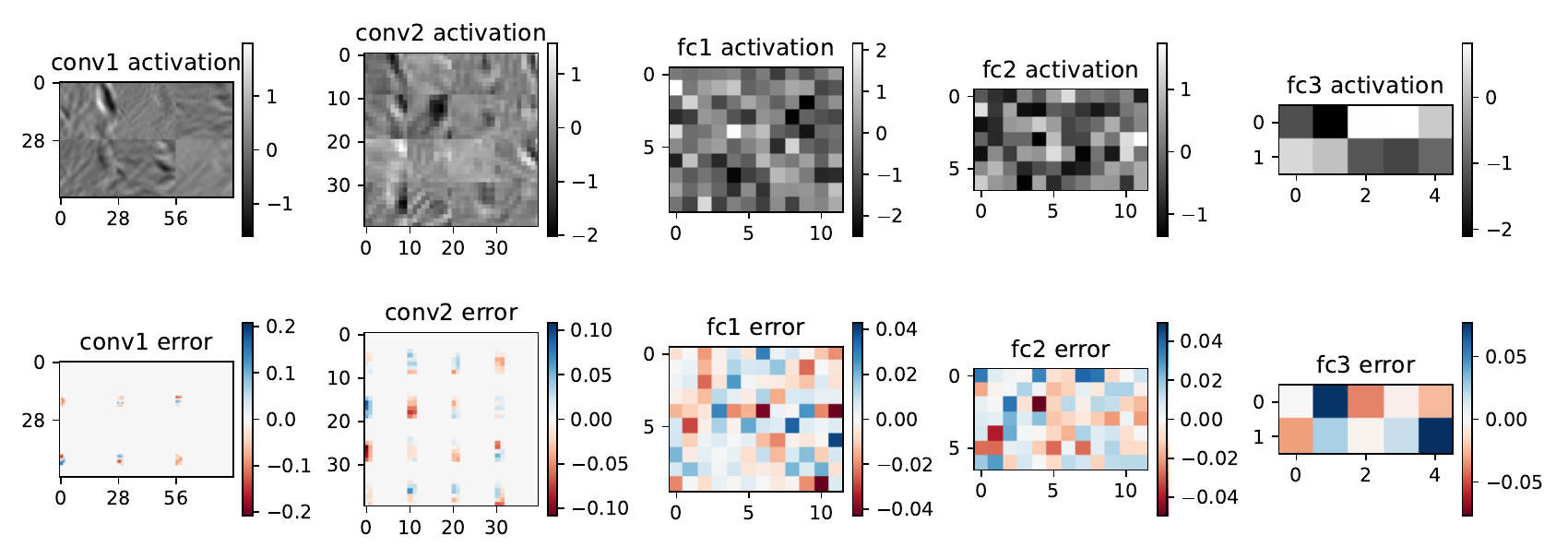}
    \caption{Visualize of LeNet internal layers and error propagation.}
    \label{fig:lenet_visualize}
\end{figure*}

Next, we evaluate error propagation using more robust metrics. Inspired by \cite{Mahmoud2020HarDNNFM}\cite{statistical2023}, MRFI provides mean absolute error (MAE) and rooted mean square error (RMSE) observers. In \autoref{fig:spread_metrics}, both MAE and RMSE reveal that weight injection has a significantly greater influence than activation value injection when performing single-point injection on the first convolution layer. This finding aligns with the accuracy results. Additionally, although the number of affected neurons rapidly increases during the propagation process, these error measures remain relatively stable.

MRFI also includes observers for recording activation and error values directly, facilitating the visualization of internal feature maps in neural networks. \autoref{fig:lenet_visualize} showcases the raw neuron values input by LeNet on a single image from the cifar-10 dataset, along with the errors caused by a single activation injection in the first layer. We observe that errors occur in fixed positions across all channels in the convolution layer's feature map, while nearly all neurons in the fully connected layers experience irregular disturbances.

\subsection{Quantization Settings Resilience Study}

Recent studies have highlighted the close relationship between quantization methods and neural network fault tolerance. Therefore, our framework integrates quantization settings into its configuration. In \autoref{sec:basic_use}, we compared the differences between two quantization methods using a simple global quantization setting. With MRFI, we can now quantize each layer of the network separately to study these differences at a fine-grained level.

\begin{figure}[htbp]
    \centering
    \includegraphics[width=0.6\linewidth]{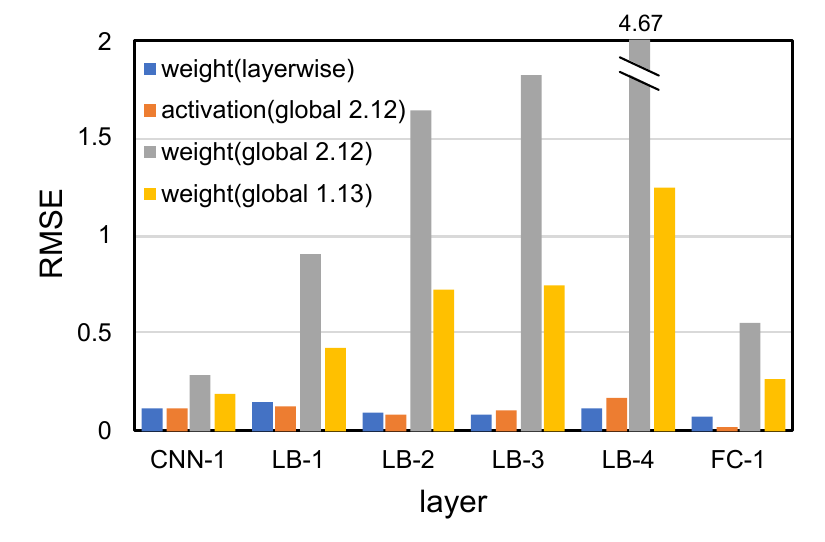}
    \caption{The difference in error sensitivity of quantization methods at different network modules.}
    \label{fig:quantization_method}
\end{figure}

\autoref{fig:quantization_method} illustrates the differences in RMSE metrics for various error injections on different layers and blocks of ResNet18. A larger RMSE value indicates greater sensitivity to errors. It is evident that global quantization settings exhibit significant variations in fault tolerance and perform poorly at each layer, resulting in lower fault tolerance than layered quantization. Conversely, layerwise quantization demonstrates stable error resilience. Additionally, we observe that weights are more sensitive to errors during global quantization, whereas weights and activation values exhibit similar sensitivities during layerwise quantization.

\begin{figure}[htbp]
    \centering
    \subfigure[Accuracy]{
        \includegraphics[width=0.45\linewidth]{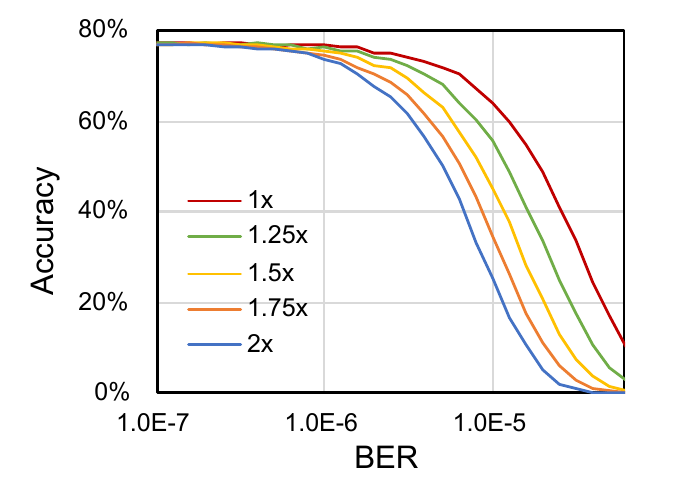}
    }
    \subfigure[RMSE]{
        \includegraphics[width=0.45\linewidth]{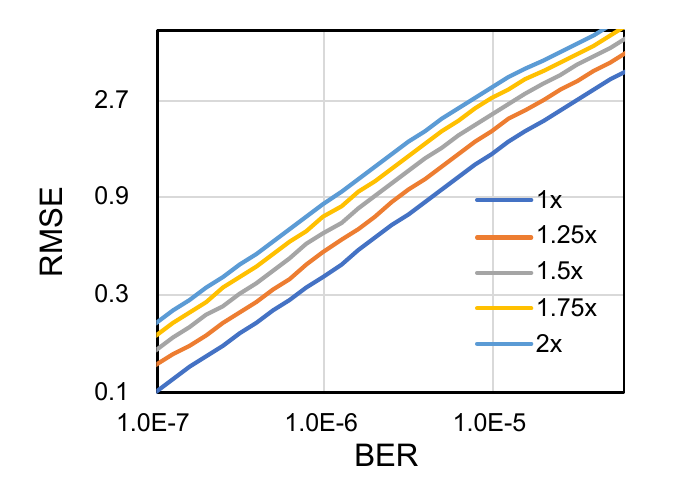}
    }
    \caption{Fault tolerance under different quantization scale parameter.}
    \label{fig:quantization_scale}
\end{figure}

The quantization parameters of the model also impact fault tolerance\cite{SNR2021}, and MRFI can easily test these settings. To demonstrate it, we employ layerwise quantization and set different quantization scaling factors for each layer of the model. Here we use a typical symmetric quantization, that is, $x_Q = clamp(round(x\times scale), Q_{min}, Q_{max})$. We evaluate accuracy and RMSE metric in ResNet as an example. As shown in \autoref{fig:quantization_scale}, larger scale factors generally lead to worse fault tolerance.

In practical deployment scenarios, models are typically quantized after training, and the quantization parameters depend on both model accuracy and target hardware requirements. The experimental differences outlined above emphasize the need for a thorough assessment of the quantization methods' impact when deploying security-critical neural network models.

\subsection{Evaluate Fault-Tolerant Resilience at Different Level}

MRFI's flexible configuration capabilities enable error injection analysis at various levels, providing robust support for selective fault-tolerant design. In this section, we demonstrate the differences in fault tolerance based on bit, channel and pixel.

\subsubsection{Bit sensitivity evaluation}

\begin{figure}[htbp]
    \centering
    \subfigure[Float16]{
        \includegraphics[width=0.45\linewidth]{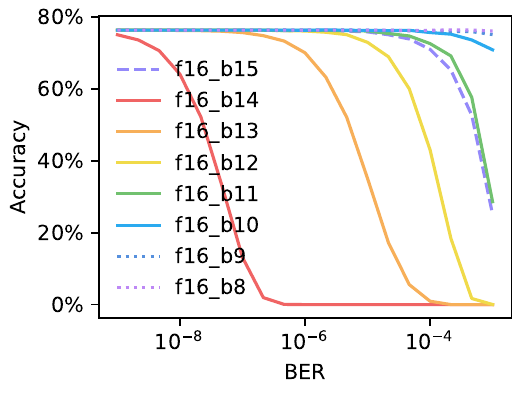}
    }
    \subfigure[Fixed 3.13 quantization]{
        \includegraphics[width=0.45\linewidth]{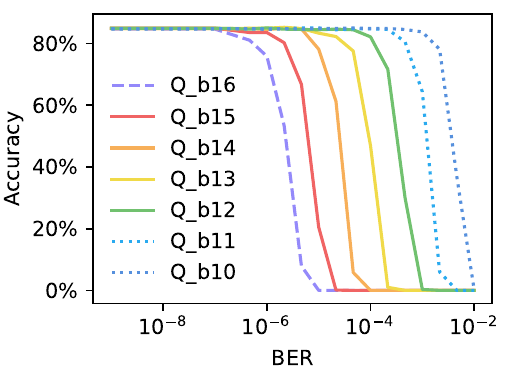}
    }
    \caption{Bit sensitivity evaluation of VGG-11 among all layers.}
    \label{fig:bit_study}
\end{figure}

We inject random errors into each bit of the float16 floating-point type and fixed-point 3.13 quantization type of the VGG-11 model, observing the model's accuracy at different injection rates. \autoref{fig:bit_study}(a) illustrates the model accuracy for the high-order bits, while the low-order bits demonstrate greater insensitivity to error injection. In float16, the highest bit represents the sign, followed by 5 exponent bits, and finally 10 mantissa bits. From \autoref{fig:bit_study}(b), we observe that high-order exponent bits are highly sensitive to errors, whereas error injection into the mantissa bits does not significantly affect the model's accuracy. The sign bit exhibits a sensitivity level lower than that of high exponent bits but higher than that of low exponent and mantissa bits. For fixed-point 3.13 quantization, a regular pattern emerges in the fault tolerance of each bit, with lower bits consistently displaying better fault tolerance than higher bits.

\subsubsection{Channel sensitivity evaluation}

\begin{figure}[htbp]
    \centering
    \includegraphics[width=0.6\linewidth]{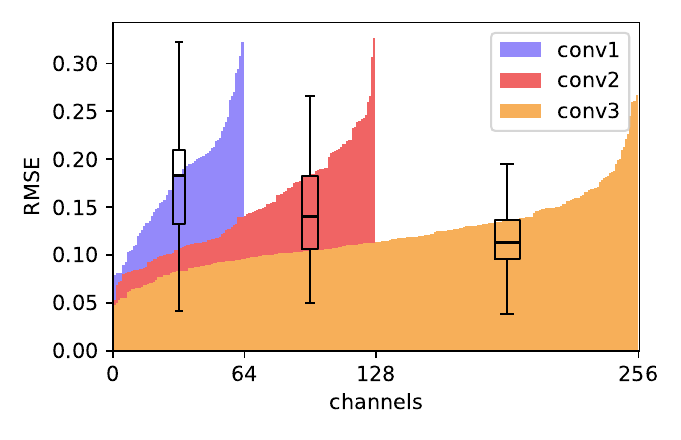}
    \caption{Channel sensitivity evaluation on three convolution layers of VGG-11. We use the RMSE metric to test the fault injection sensitivity of each channel and sort them in ascending order. The box plot shows the median and quantile of the data, indicating significant differences between channels.}
    \label{fig:channel_wise}
\end{figure}

\autoref{fig:channel_wise} showcases the differences in sensitivity indicators for each channel in the activation values of the first three layers of VGG-11. The three convolutional layers contain 64, 128, and 256 feature map channels, respectively. By employing layerwise quantization parameters, the differences in fault tolerance among different layers are small. After sorting the RMSE indicators for each channel in ascending order, we observe that the variance of channel sensitivity within a layer surpasses the variance of sensitivity between layers. This suggests that selective fault-tolerant design at the channel level can yield enhanced fault tolerance outcomes.

\subsubsection{Pixel sensitivity evaluation}

\begin{figure}[htbp]
    \centering
    \includegraphics[width=0.9\linewidth]{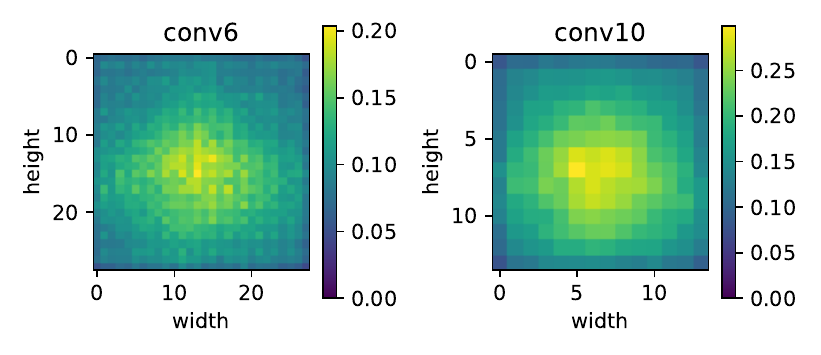}
    \caption{Pixel sensitivity evaluation on two convolution layers of ResNet18 using RMSE metric. The data for each pixel comes from random bit injection along all channels of the feature map.}
    \label{fig:pixel_wise}
\end{figure}

It is also possible to evaluate the pixel sensitivity differences of feature maps at the spatial level. We use MRFI to randomly inject faults into each channel of different pixels in two layers of ResNet, resulting in the output RMSE as shown in \autoref{fig:pixel_wise}. We can see that pixels in the center position are usually more sensitive to errors, which may be related to the dataset features of ImageNet, as most targets are located at the center of the image.

\subsection{Performance}

In addition to accuracy, the performance of error injection and simulation in neural networks is a crucial factor for conducting experiments. MRFI has implemented optimizations in error injection, resulting in comparable time costs between error injection and the golden run. Below, we will refer the state of the art tools PyTorchFI and Ares to conduct two similar experiments and measure the time consumption. PyTorchFI has been optimized for single point error injection, making its overhead negligible compared to forward propagation. However, PyTorchFI does not provide random error generation based on injection rate, so we compared it with the random sampling method used by Ares as the baseline.

\subsubsection{Single point injection}
\autoref{tab:perf} presents the evaluation time for single point injection errors on GPUs and CPUs. It demonstrates that the execution time of the MRFI framework can be disregarded relative to the model's computation. This indicates that although our tool is designed for complex injection configurations, it has also achieved target performance close to PyTorchFI.

\subsubsection{Injection according to bit error rate}
\begin{figure}[htbp]
    \centering
    \subfigure[Time cost]{
        \includegraphics[width=0.45\linewidth]{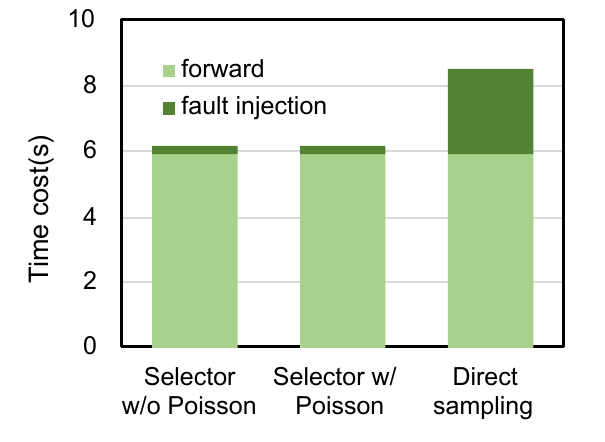}
    }
    \subfigure[Accuracy]{
        \includegraphics[width=0.45\linewidth]{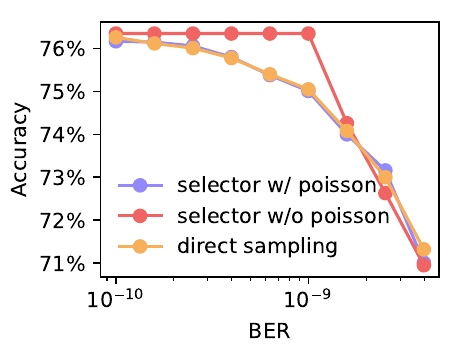}
    }
    \caption{Comparison of performance and accuracy of different random sampling methods for random multi-point injection.}
    \label{fig:poisson_perf}
\end{figure}

For random multi-point injection, Ares conducts acceptance-rejection sampling on each bit, while MRFI uses selectors to choose injection error positions based on the expected number calculated using \autoref{eq:inject_naive}. This reduces the number of random number generations required. However, a discrepancy arises when the injection error rate is low, as $n_{inject}$ is a discrete integer. To address this issue, we introduce a Poisson sampling strategy described in \autoref{eq:inject_poisson}, ensuring consistent results in theory. \autoref{fig:poisson_perf} compares the performance and accuracy between direct probability sampling and optimization methods. While using a integer selector can significantly reduce the cost of sampling calculations, it may introduce jumps at low injection rates. On the other hand, Poisson sampling maintains consistent results with the baseline sampling methodology.

\begin{equation}
    n_{inject} = \lfloor N_{neurons} \times BER \rceil
    \label{eq:inject_naive}
\end{equation}

\begin{equation}
    N_{inject} \sim Poisson(N_{neurons} \times BER)
    \label{eq:inject_poisson}
\end{equation}

\begin{table}[htbp]
  \centering
  \caption{Time cost comparison on single point injection}
    \begin{tabular}{|l|r|l|l|}
    \hline
      & \multicolumn{1}{l|}{Acc.} & GPU time & CPU time \\
    \hline
    golden run & 68.87\% & 86.83±0.06s & 2603±2s \\
    \hline
    1p fault inject & 68.33\% & 87.94±0.04s & 2607±3s \\
    \hline
    \end{tabular}%
  \label{tab:perf}%
\end{table}%

\subsection{Comparison to Fault Simulation with Architecture Details}

\begin{figure}[htbp]
    \centering
    \subfigure[Percentage of classification errors caused by each bit]{
        \includegraphics[width=0.75\linewidth]{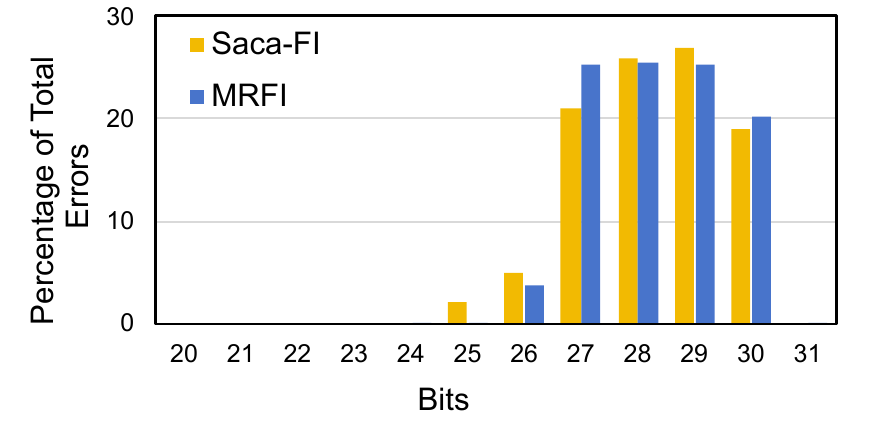}
    }
    \subfigure[Accuracy on different number of fault points]{
        \includegraphics[width=0.75\linewidth]{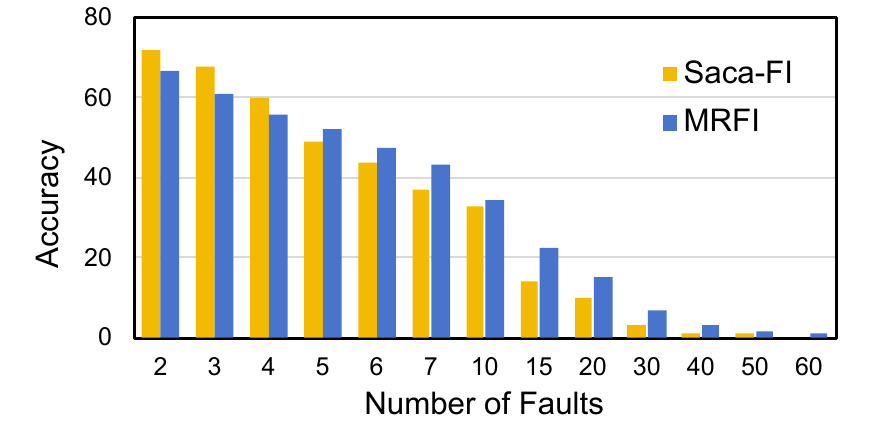}
    }
    \caption{Comparison between MRFI and fault injection tool with microarchitecture simulation Saca-FI\cite{tan2023saca} on VGG-16 network with float32 data type. MRFI and Saca-FI showed consistent trends in the results of single point injection experiments and multi-point injection experiments.}
    \label{fig:saca}
\end{figure}

Since bit-flip errors essentially occur in hardware, fault injection to simulators with architectural details ensures more accurate evaluation of the influence of hardware faults on neural network processing. Despite the fault simulation accuracy, the fault simulation can be limited to specific hardware architectures which makes the analysis be lack of generality. Moreover, cycle accurate simulation can be extremely slow and prohibitively expensive because evaluating influence of hardware errors on the model accuracy requires inference of a large number of inputs. In order to demonstrate the accuracy of the proposed fault simulation framework MRFI, we take Saca-FI \cite{tan2023saca}, which conducts fault injection and analysis over a neural network accelerator simulator with architecture details, as a golden reference. The fault injection is performed on a VGG-16 model with 32-bit float data type. In \autoref{fig:saca}(a), we inject single bit flip to various bit locations of a random activation and analyze the percentage of classification errors induced by the bit flip error. It demonstrates that the results of Saca-FI and MRFI are generally consistent and the majority of classification errors are mostly caused by high exponent bits of the floating numbers. For multiple bit flip errors as shown in \autoref{fig:saca}(b), the experiments show the model accuracy under different number of bit-flip errors. It can be seen that MRFI achieves similar accuracy degradation trend as Saca-FI in general. The major difference can be attributed to the architecture-induced computing such as data flow and compilation. In contrast, MRFI treats all the computing and data equally. 

\section{Conclusion}
In this paper, we have introduced MRFI, an open-source multi-resolution fault injection framework for neural network processing. MRFI enables refined error assessments, considering the impact of model quantization and supporting fine-grained error injection configurations. To simplify configuration, MRFI decompose complex configuration requirements into multiple modules and offers a user-friendly configuration interface. Our experiments have shown that MRFI can evaluate errors from multiple aspects with practical needs.




\bibliographystyle{IEEEtran}
\bibliography{cite}

\begin{thebibliography}{10}
\providecommand{\url}[1]{#1}
\csname url@samestyle\endcsname
\providecommand{\newblock}{\relax}
\providecommand{\bibinfo}[2]{#2}
\providecommand{\BIBentrySTDinterwordspacing}{\spaceskip=0pt\relax}
\providecommand{\BIBentryALTinterwordstretchfactor}{4}
\providecommand{\BIBentryALTinterwordspacing}{\spaceskip=\fontdimen2\font plus
\BIBentryALTinterwordstretchfactor\fontdimen3\font minus
  \fontdimen4\font\relax}
\providecommand{\BIBforeignlanguage}[2]{{%
\expandafter\ifx\csname l@#1\endcsname\relax
\typeout{** WARNING: IEEEtran.bst: No hyphenation pattern has been}%
\typeout{** loaded for the language `#1'. Using the pattern for}%
\typeout{** the default language instead.}%
\else
\language=\csname l@#1\endcsname
\fi
#2}}
\providecommand{\BIBdecl}{\relax}
\BIBdecl

\bibitem{mittal2020survey}
S.~Mittal, ``A survey on modeling and improving reliability of dnn algorithms
  and accelerators,'' \emph{Journal of Systems Architecture}, vol. 104, p.
  101689, 2020.

\bibitem{ibrahim2020soft}
Y.~Ibrahim, H.~Wang, J.~Liu, J.~Wei, L.~Chen, P.~Rech, K.~Adam, and G.~Guo,
  ``Soft errors in dnn accelerators: A comprehensive review,''
  \emph{Microelectronics Reliability}, vol. 115, p. 113969, 2020.

\bibitem{wei2023tc}
X.~Wei, C.~Zhou, H.~Yue, and J.~T. Zhou, ``Tc-sepm: Characterizing soft error
  resilience of cnns on tensor cores from program and microarchitecture
  perspectives,'' \emph{Journal of Systems Architecture}, vol. 145, p. 103024,
  2023.

\bibitem{nguyen2023craft}
T.-H. Nguyen, M.~Imran, J.~Choi, and J.-S. Yang, ``Craft: Criticality-aware
  fault-tolerance enhancement techniques for emerging memories-based deep
  neural networks,'' \emph{IEEE Transactions on Computer-Aided Design of
  Integrated Circuits and Systems}, 2023.

\bibitem{wang2023reliable}
Y.~Wang, S.~Shi, X.~He, Z.~Tang, X.~Pan, Y.~Zheng, X.~Wu, A.~C. Zhou, B.~He,
  and X.~Chu, ``Reliable and efficient in-memory fault tolerance of large
  language model pretraining,'' \emph{arXiv preprint arXiv:2310.12670}, 2023.

\bibitem{he2023understanding}
Y.~He, M.~Hutton, S.~Chan, R.~De~Gruijl, R.~Govindaraju, N.~Patil, and Y.~Li,
  ``Understanding and mitigating hardware failures in deep learning training
  systems,'' in \emph{Proceedings of the 50th Annual International Symposium on
  Computer Architecture}, 2023, pp. 1--16.

\bibitem{wu2023transom}
B.~Wu, L.~Xia, Q.~Li, K.~Li, X.~Chen, Y.~Guo, T.~Xiang, Y.~Chen, and S.~Li,
  ``Transom: An efficient fault-tolerant system for training llms,''
  \emph{arXiv preprint arXiv:2310.10046}, 2023.

\bibitem{PytorchFI}
A.~Mahmoud, N.~Aggarwal, A.~Nobbe, J.~R.~S. Vicarte, S.~V. Adve, C.~W.
  Fletcher, I.~Frosio, and S.~K.~S. Hari, ``{PyTorchFI: A Runtime Perturbation
  Tool for DNNs},'' \emph{Proceedings - 50th Annual IEEE/IFIP International
  Conference on Dependable Systems and Networks, DSN-W 2020}, pp. 25--31, 2020.

\bibitem{Ares2018}
B.~Reagen, U.~Gupta, L.~Pentecost, P.~Whatmough, S.~K. Lee, N.~Mulholland,
  D.~Brooks, and G.~Y. Wei, ``{Ares: A framework for quantifying the resilience
  of deep neural networks},'' in \emph{Proceedings - Design Automation
  Conference}, vol. Part F1377, 2018.

\bibitem{TensorFI}
Z.~Chen, N.~Narayanan, B.~Fang, G.~Li, K.~Pattabiraman, and N.~DeBardeleben,
  ``{Tensorfi: A flexible fault injection framework for tensorflow
  applications},'' \emph{Proceedings - International Symposium on Software
  Reliability Engineering, ISSRE}, vol. 2020-Octob, pp. 426--435, 2020.

\bibitem{torchfi2019}
B.~Goldstein, S.~Srinivasan, N.~K. Mellempudi, D.~Das, L.~Santiago, V.~C.
  Ferreira, N.~Solon, S.~Kundu, and F.~M.~G. França, ``Reliability evaluation
  of compressed deep learning models,'' in \emph{2020 IEEE 11th Latin American
  Symposium on Circuits Systems (LASCAS)}, 2020.

\bibitem{Zhen2021MindFI}
Y.~Zheng, Z.~Feng, Z.~Hu, and K.~Pei, ``Mindfi: A fault injection tool for
  reliability assessment of mindspore applicacions,'' in \emph{2021 IEEE
  International Symposium on Software Reliability Engineering Workshops
  (ISSREW)}, 2021, pp. 235--238.

\bibitem{xue2023exploring}
X.~Xue, C.~Liu, B.~Liu, H.~Huang, Y.~Wang, T.~Luo, L.~Zhang, H.~Li, and X.~Li,
  ``Exploring winograd convolution for cost-effective neural network fault
  tolerance,'' \emph{IEEE Transactions on Very Large Scale Integration (VLSI)
  Systems}, 2023.

\bibitem{xue2023soft}
X.~Xue, C.~Liu, Y.~Wang, B.~Yang, T.~Luo, L.~Zhang, H.~Li, and X.~Li, ``Soft
  error reliability analysis of vision transformers,'' \emph{IEEE Transactions
  on Very Large Scale Integration (VLSI) Systems}, 2023.

\bibitem{Mahmoud2021ISSRE}
A.~Mahmoud, S.~K. Sastry~Hari, C.~W. Fletcher, S.~V. Adve, C.~Sakr,
  N.~Shanbhag, P.~Molchanov, M.~B. Sullivan, T.~Tsai, and S.~W. Keckler,
  ``Optimizing selective protection for cnn resilience,'' in \emph{2021 IEEE
  32nd International Symposium on Software Reliability Engineering (ISSRE)},
  2021, pp. 127--138.

\bibitem{statistical2023}
H.~Huang, X.~Xue, C.~Liu, Y.~Wang, T.~Luo, L.~Cheng, H.~Li, and X.~Li,
  ``Statistical modeling of soft error influence on neural networks,''
  \emph{IEEE Transactions on Computer-Aided Design of Integrated Circuits and
  Systems}, pp. 1--1, 2023.

\bibitem{FIdelity}
Y.~He, P.~Balaprakash, and Y.~Li, ``{Fidelity: Efficient resilience analysis
  framework for deep learning accelerators},'' \emph{Proceedings of the Annual
  International Symposium on Microarchitecture, MICRO}, vol. 2020-Octob, pp.
  270--281, 2020.

\bibitem{libano2018selective}
F.~Libano, B.~Wilson, J.~Anderson, M.~J. Wirthlin, C.~Cazzaniga, C.~Frost, and
  P.~Rech, ``Selective hardening for neural networks in fpgas,'' \emph{IEEE
  Transactions on Nuclear Science}, vol.~66, no.~1, pp. 216--222, 2018.

\bibitem{ahmadilivani2023enhancing}
M.~H. Ahmadilivani, M.~Taheri, J.~Raik, M.~Daneshtalab, and M.~Jenihhin,
  ``Enhancing fault resilience of qnns by selective neuron splitting,''
  \emph{arXiv preprint arXiv:2306.09973}, 2023.

\bibitem{SNR2021}
\BIBentryALTinterwordspacing
E.~Ozen and A.~Orailoglu, ``{SNR: Squeezing Numerical Range Defuses Bit Error
  Vulnerability Surface in Deep Neural Networks},'' \emph{ACM Trans. Embed.
  Comput. Syst.}, vol.~20, no.~5s, 2021. [Online]. Available:
  \url{https://doi.org/10.1145/3477007}
\BIBentrySTDinterwordspacing

\bibitem{SASSIFI}
S.~K.~S. Hari, T.~Tsai, M.~Stephenson, S.~W. Keckler, and J.~Emer, ``{SASSIFI:
  An architecture-level fault injection tool for GPU application resilience
  evaluation},'' \emph{ISPASS 2017 - IEEE International Symposium on
  Performance Analysis of Systems and Software}, vol.~1, no.~1, pp. 249--258,
  2017.

\bibitem{tsai2021nvbitfi}
T.~Tsai, S.~K.~S. Hari, M.~Sullivan, O.~Villa, and S.~W. Keckler, ``Nvbitfi:
  Dynamic fault injection for gpus,'' in \emph{2021 51st Annual IEEE/IFIP
  International Conference on Dependable Systems and Networks (DSN)}.\hskip 1em
  plus 0.5em minus 0.4em\relax IEEE, 2021, pp. 284--291.

\bibitem{gambardella2019efficient}
G.~Gambardella, J.~Kappauf, M.~Blott, C.~Doehring, M.~Kumm, P.~Zipf, and
  K.~Vissers, ``Efficient error-tolerant quantized neural network
  accelerators,'' in \emph{2019 IEEE International Symposium on Defect and
  Fault Tolerance in VLSI and Nanotechnology Systems (DFT)}.\hskip 1em plus
  0.5em minus 0.4em\relax IEEE, 2019, pp. 1--6.

\bibitem{xu2020persistent}
D.~Xu, Z.~Zhu, C.~Liu, Y.~Wang, H.~Li, L.~Zhang, and K.-T. Cheng, ``Persistent
  fault analysis of neural networks on fpga-based acceleration system,'' in
  \emph{2020 IEEE 31st International Conference on Application-specific
  Systems, Architectures and Processors (ASAP)}.\hskip 1em plus 0.5em minus
  0.4em\relax IEEE, 2020, pp. 85--92.

\bibitem{xu2021reliability}
D.~Xu, Z.~Zhu, C.~Liu, Y.~Wang, S.~Zhao, L.~Zhang, H.~Liang, H.~Li, and K.-T.
  Cheng, ``Reliability evaluation and analysis of fpga-based neural network
  acceleration system,'' \emph{IEEE Transactions on Very Large Scale
  Integration (VLSI) Systems}, vol.~29, no.~3, pp. 472--484, 2021.

\bibitem{tan2023saca}
J.~Tan, Q.~Wang, K.~Yan, X.~Wei, and X.~Fu, ``Saca-fi: A
  microarchitecture-level fault injection framework for reliability analysis of
  systolic array based cnn accelerator,'' \emph{Future Generation Computer
  Systems}, vol. 147, pp. 251--264, 2023.

\bibitem{Li2017understanding}
\BIBentryALTinterwordspacing
G.~Li, S.~K.~S. Hari, M.~B. Sullivan, T.~Tsai, K.~Pattabiraman, J.~S. Emer, and
  S.~W. Keckler, ``Understanding error propagation in deep learning neural
  network {(DNN)} accelerators and applications,'' in \emph{Proceedings of the
  International Conference for High Performance Computing, Networking, Storage
  and Analysis, {SC} 2017, Denver, CO, USA, November 12 - 17, 2017}, B.~Mohr
  and P.~Raghavan, Eds.\hskip 1em plus 0.5em minus 0.4em\relax {ACM}, 2017, pp.
  8:1--8:12. [Online]. Available: \url{https://doi.org/10.1145/3126908.3126964}
\BIBentrySTDinterwordspacing

\bibitem{xue2022winograd}
X.~Xue, H.~Huang, C.~Liu, T.~Luo, L.~Zhang, and Y.~Wang, ``Winograd
  convolution: A perspective from fault tolerance,'' in \emph{Proceedings of
  the 59th ACM/IEEE Design Automation Conference}, 2022, pp. 853--858.

\bibitem{Mahmoud2020HarDNNFM}
A.~Mahmoud, S.~K.~S. Hari, C.~W. Fletcher, S.~V. Adve, C.~Sakr, N.~R. Shanbhag,
  P.~Molchanov, M.~B. Sullivan, T.~Tsai, and S.~W. Keckler, ``Hardnn: Feature
  map vulnerability evaluation in cnns,'' \emph{ArXiv}, vol. abs/2002.09786,
  2020.

\bibitem{chen2019BinFI}
\BIBentryALTinterwordspacing
Z.~Chen, G.~Li, K.~Pattabiraman, and N.~DeBardeleben, ``Binfi: An efficient
  fault injector for safety-critical machine learning systems,'' in
  \emph{Proceedings of the International Conference for High Performance
  Computing, Networking, Storage and Analysis}, ser. SC '19.\hskip 1em plus
  0.5em minus 0.4em\relax New York, NY, USA: Association for Computing
  Machinery, 2019. [Online]. Available:
  \url{https://doi.org/10.1145/3295500.3356177}
\BIBentrySTDinterwordspacing

\bibitem{liu2022special}
C.~Liu, Z.~Gao, S.~Liu, X.~Ning, H.~Li, and X.~Li, ``Special session:
  Fault-tolerant deep learning: A hierarchical perspective,'' in \emph{2022
  IEEE 40th VLSI Test Symposium (VTS)}.\hskip 1em plus 0.5em minus 0.4em\relax
  IEEE, 2022, pp. 1--12.

\bibitem{li2020ftt}
W.~Li, X.~Ning, G.~Ge, X.~Chen, Y.~Wang, and H.~Yang, ``Ftt-nas: Discovering
  fault-tolerant neural architecture,'' in \emph{2020 25th Asia and South
  Pacific Design Automation Conference (ASP-DAC)}.\hskip 1em plus 0.5em minus
  0.4em\relax IEEE, 2020, pp. 211--216.

\bibitem{ning2021ftt}
X.~Ning, G.~Ge, W.~Li, Z.~Zhu, Y.~Zheng, X.~Chen, Z.~Gao, Y.~Wang, and H.~Yang,
  ``Ftt-nas: Discovering fault-tolerant convolutional neural architecture,''
  \emph{ACM Transactions on Design Automation of Electronic Systems (TODAES)},
  vol.~26, no.~6, pp. 1--24, 2021.

\bibitem{zhang2018analyzing}
J.~J. Zhang, T.~Gu, K.~Basu, and S.~Garg, ``Analyzing and mitigating the impact
  of permanent faults on a systolic array based neural network accelerator,''
  in \emph{2018 IEEE 36th VLSI Test Symposium (VTS)}.\hskip 1em plus 0.5em
  minus 0.4em\relax IEEE, 2018, pp. 1--6.

\bibitem{reagen2016minerva}
B.~Reagen, P.~Whatmough, R.~Adolf, S.~Rama, H.~Lee, S.~K. Lee, J.~M.
  Hern{\'a}ndez-Lobato, G.-Y. Wei, and D.~Brooks, ``Minerva: Enabling
  low-power, highly-accurate deep neural network accelerators,'' in \emph{2016
  ACM/IEEE 43rd Annual International Symposium on Computer Architecture
  (ISCA)}.\hskip 1em plus 0.5em minus 0.4em\relax IEEE, 2016, pp. 267--278.

\bibitem{hari22022abft}
S.~K.~S. Hari, M.~B. Sullivan, T.~Tsai, and S.~W. Keckler, ``Making
  convolutions resilient via algorithm-based error detection techniques,''
  \emph{IEEE Transactions on Dependable and Secure Computing}, vol.~19, no.~4,
  pp. 2546--2558, 2022.

\bibitem{fitact2022}
B.~Ghavami, M.~Sadati, Z.~Fang, and L.~Shannon, ``Fitact: Error resilient deep
  neural networks via fine-grained post-trainable activation functions,'' in
  \emph{2022 Design, Automation \& Test in Europe Conference \& Exhibition
  (DATE)}, 2022, pp. 1239--1244.

\bibitem{xu2021r2f}
D.~Xu, M.~He, C.~Liu, Y.~Wang, L.~Cheng, H.~Li, X.~Li, and K.-T. Cheng, ``R2f:
  A remote retraining framework for aiot processors with computing errors,''
  \emph{IEEE Transactions on Very Large Scale Integration (VLSI) Systems},
  vol.~29, no.~11, pp. 1955--1966, 2021.

\end{thebibliography}

\end{document}